\documentclass[11pt,a4paper]{article}
\usepackage[hyperref]{naaclhlt2019}
\usepackage{times}
\usepackage{latexsym}
\usepackage{graphicx}  
\usepackage{float}
\usepackage{multirow}
\usepackage{url}

\aclfinalcopy 

\title{IMHO Fine-Tuning Improves Claim Detection}

\author{Tuhin Chakrabarty \\
  Dept. Of Computer Science \\
  Columbia University \\
  {\tt tc2896@columbia.edu} \\\And
  Christopher Hidey \\
  Dept. Of Computer Science \\
  Columbia University \\
  {\tt ch3085@columbia.edu} \
  \\\And
  Kathleen McKeown \\
  Dept. Of Computer Science \\
  Columbia University \\
  {\tt kathy@cs.columbia.edu} \\}

\date{}

\begin{document}
\maketitle
\begin{abstract}
Claims are the central component of an argument. Detecting claims across different domains or data sets can often be challenging due to their varying conceptualization. We propose to alleviate this problem by fine tuning  a language model using a Reddit corpus of 5.5  million opinionated claims. These claims are self-labeled by their authors using the internet acronyms IMO/IMHO (in my (humble) opinion). Empirical results show that using this approach improves the state of art performance across four benchmark argumentation data sets by an average of 4 absolute F1 points in claim detection. As these data sets include diverse domains such as social media and student essays this improvement demonstrates the robustness of fine-tuning on this novel corpus. 
\end{abstract}

\section{Introduction}


Toulmin's influential work on argumentation \shortcite{Toulmin} introduced a claim as an \textit{assertion that deserves our attention}. More recent work describes a claim as \textit{a statement that is in dispute and that we are trying to support with reasons} \cite{Govier}. 
While some traits of claims are defined by their context, such as that claims usually need some support to
make up a 'complete' argument (e.g., premises, evidence,
or justifications), the exact definition of a claim may vary depending on the domain, register, or task. \newcite{essence} try to solve the problem of claim conceptualization by training models across one data set and testing on others, but their cross-domain claim detection experiments mostly led to decreased results over in-domain experiments.

To demonstrate that some properties of claims are shared across domains, we create a diverse and rich corpus mined from Reddit and evaluate on held out datasets from different sources. 
We use Universal Language Model Fine-Tuning (ULMFiT) \cite{ulmfit},  which pre-trains a language model (LM) on a large general-domain corpus and fine-tunes it on our Reddit corpus before training a final classifier to identify claims on various data sets.

We make the following contributions:  
\begin{itemize}
	\item We release a dataset of 5.5 million opinionated claims from Reddit,\footnote{https://bitbucket.org/tuhinch/imho-naacl2019} which we hope will be useful for computational argumentation.
	\item We show transfer learning helps in the detection of claims with varying definitions and conceptualizations across data sets from diverse domains such as social media and student essays.
	\item Empirical results show that using the Reddit corpus for language model fine-tuning improves the state-of-the-art performance across four benchmark argumentation data sets by an average of 4 absolute F1 points in claim detection. 
\end{itemize}


\section{Related Work}
Argumentation mining (AM) is a research field within the growing area of computational argumentation. The tasks pursued within this field are highly challenging and include segmenting argumentative and non-argumentative text units, parsing argument structures, and recognizing argumentative components such as claims-— the main focus of this work. On the modeling side, \newcite{pe} and \newcite{essay} used pipeline approaches for AM, combining parts of the pipeline using integer linear programming (ILP). \newcite{end2end} proposed state-of-art sequence tagging neural end-to-end models for AM. \newcite{multi} used multi-task learning (MTL) to identify argumentative components, challenging assumptions that conceptualizations across AM data sets are divergent and that MTL is difficult for semantic or higher-level tasks.

\newcite{sara} were among the first to conduct cross-domain experiments for claim detection. However they focused on relatively similar data sets like blog articles from LiveJournal and Wikipedia discussions. \newcite{khatib}, on the other hand, wanted to identify argumentative sentences through cross-domain experiments. Their goal was, however, to improve argumentation mining via distant supervision
rather than detecting differences in the notions of a claim. \newcite{essence} showed that while the divergent conceptualization of claims in different data sets is indeed harmful to cross-domain classification, there are shared properties on the lexical level as well as system configurations that can help to overcome these gaps. To this end they carried out  experiments using models with engineered features and deep learning to identify claims in a cross-domain fashion.

Pre-trained language models have been recently used to achieve state-of-the-art results on a wide range of NLP tasks (e.g., sequence labeling and sentence classification). Some of the recent works that have employed pre-trained language models include ULMFiT \cite{ulmfit}, ELMo \cite{Peters:2018}, GLoMo \cite{glomo}, BERT \cite{bert} and OpenAI transformer \cite{openai}. While these models have demonstrated success on a variety of tasks, they have yet to be widely used in argumentation mining. 

\section{Data}
As the goal of our experiments is to develop models that generalize across domains, we collect a large, diverse dataset from social media and fine-tune and evaluate on held out data sets.

\begin{figure*}[h]
\centering
\includegraphics[scale=0.75, trim={0 1cm 0 0},]{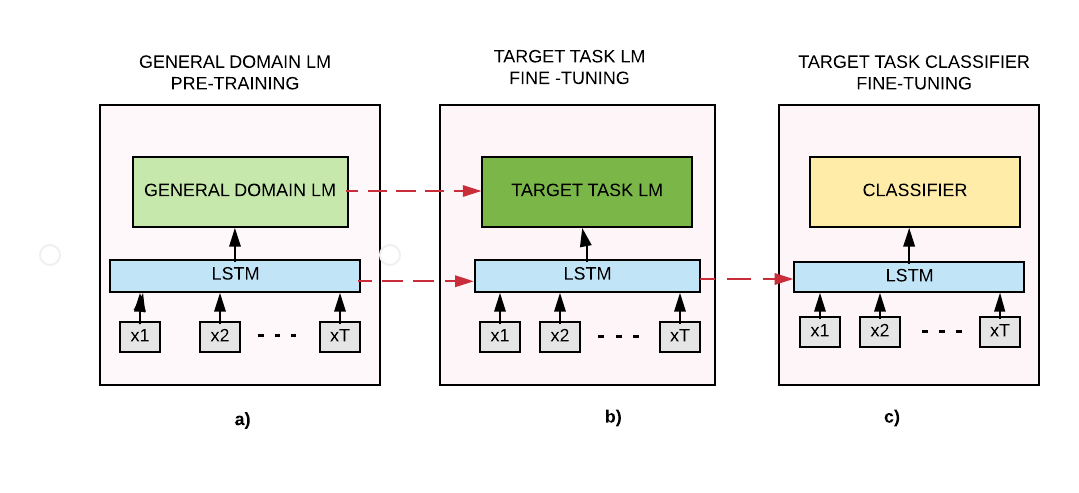}
\caption{\label{modelpic}Schematic of ULMFiT, showing three stages.  The dashed arrows indicate that the parameters from the previous stage were used to initialize the next stage.}
\label{fig:fig2}
\end{figure*}

\subsection{Self-labeled Opinion Data Collection}
In order to obtain a data set representative of claims, we need a method of automatic data collection that 
introduces minimal linguistic bias.  We thus mine comments containing the acronyms IMO (in my opinion) or IMHO (in my humble opinion) from the social media site Reddit. IM(H)O is a commonly used acronym\footnote{https://reddit.zendesk.com/hc/en-us/articles/205173295-What-do-all-these-acronyms-mean-} with the only purpose of identifying one's own comment as a personal opinion.  We provide some examples\footnote{Examples have been modified to protect user privacy} below:
\begin{quote}
That's virtually the same as neglect right there \textbf{IMHO}.
\end{quote}
\begin{quote}
\textbf{IMO}, Lakers are in big trouble next couple years
\end{quote}
To use these examples for pre-training, we need only to remove the acronym (and any resulting unnecessary punctuation).

We collect Reddit comments from December 2008 to August 2017 through the pushshift.io API, resulting in 5,569,962 comments.  We perform sentence and word tokenization using Spacy.  We then extract only the sentence containing IMO or IMHO and discarded the surrounding text. We refer to the resulting collection of comments as the \textbf{IMHO} dataset. 

\subsection{Labeled Claim Data}
The IMHO dataset contains no negative examples, only labeled opinions.  Furthermore, opinions in this dataset may be only a claim or both a claim and a premise. 
As our goal is to identify claims, 
we thus consider four data sets from argumentation mining.  As argumentation appears in both monologue and dialogue data, we choose two datasets created from student essays and two from social media.
\newcite{micro} created a corpus of German microtexts (\textbf{MT}) of controlled linguistic and rhetorical complexity. Each document includes a single argument and does not exceed five argumentative components.  This corpus was translated to English, which we use for our experiments. The persuasive essay (\textbf{PE}) corpus \cite{pe} includes 402 student essays. The scheme comprises major claims, claims, and premises at the clause level. This corpus has been used extensively in the argumentation mining community. The corpus from \newcite{wd} includes user-generated web discourse (\textbf{WD}) such as blog posts, or user comments annotated with claims and premises as well as backings, rebuttals and refutations. Finally, \newcite{cmv} propose a two-tiered annotation scheme to label claims and premises and their semantic types in an online persuasive forum (\textbf{CMV}) using a sample of 78 threads from the sub-reddit Change My View, with the long-term goal of understanding what makes a message persuasive.
As with \newcite{essence}, we model claim detection at the sentence level, as this
is the only way to make all data sets compatible to
each other. Table \ref{datasets} gives an overview of the data.

\begin{table}[]
\small
\begin{tabular}{|c|c|c|c|}
\hline
 & \#Claims & \#Sentences & \%Claims \\ \hline
\textbf{MT} & 112 & 449 & 24.94 \\ \hline
\textbf{PE} & 2108 & 7116 & 29.62 \\ \hline
\textbf{WD} & 211 & 3899 & 5.41 \\ \hline
\textbf{CMV} & 1206 & 3541 & 34.0 \\ \hline
\end{tabular}
\caption{\label{datasets} Table showing number of claims and total number of sentences in the data sets along with the percentage of claims in them}
\end{table}

\section{Model}

As the IMHO dataset is only self-labeled with claim data but does not contain non-claims, we need a method of incorporating this dataset into a claim detection model.
We thus use a language model fine-tuning approach, which requires only data similar to the task of interest.

The Universal Language Model Fine-Tuning method (ULMFiT) \cite{ulmfit} consists of the following steps: a) General-domain LM pre-training  b) Task-specific LM fine-tuning and c) Task-specific classifier fine-tuning. In step (a), the language model is trained on Wikitext-103 \cite{merity} consisting of 28,595 preprocessed Wikipedia articles and 103 million words capturing general properties of language. Step (b) fine-tunes the LM on task-specific data, as no matter how diverse the general-domain data used for pre-training is, the data of the target task will likely come from a different distribution.  In step (c), a classifier is then trained on the target task, fine-tuning the pre-trained LM but with an additional layer for class prediction. 
The models all use a stacked LSTM to represent each sentence.  For stages (a) and (b), the output of the LSTM is used to make a prediction of the next token and the parameters from stage (a) are used to initialize stage (b).  For stage (c), the model is initialized with the same LSTM but with a new classifier layer given the output of the LSTM.

This process is illustrated in Figure \ref{fig:fig2}.
We refer the reader to \newcite{ulmfit} for further details.

In our work, we maintain steps (a) and (c) but modify step (b) so that we fine-tune the language model on our IMHO dataset rather than the task-specific data. The goal of ULMFiT is to allow training on small datasets of only a few hundred examples,
but our experiments will show that fine-tuning the language model on opinionated claims improves over only task-specific LM fine-tuning.

\begin{table*}[]
\center
\begin{tabular}{|c|c|c|c|c|c|c|c|}
\hline
\multirow{2}{*}{} & \multirow{2}{*}{Metric} & \multicolumn{2}{l|}{CNN} & \multicolumn{2}{l|}{\begin{tabular}[c]{@{}l@{}}Task-Specific \\LM Fine-Tuning\end{tabular}} & \multicolumn{2}{l|}{\begin{tabular}[c]{@{}l@{}}IMHO  LM\\ Fine-Tuning\end{tabular}} \\ \cline{3-8} 
 &  & Claim & Macro & Claim & Macro & Claim & Macro \\ \hline
\multirow{3}{*}{WD} & P & 50.0 & 72.5 & 50.0 & 72.5 & \textbf{54.0} & \textbf{75.9} \\ \cline{2-8} 
 & R & 20.4 & 59.2 & 20.0 & 59.8 & \textbf{24.0} & \textbf{61.7} \\ \cline{2-8} 
 & F & 28.9 & 62.6 & 28.5 & 62.7 & \textbf{33.3} & \textbf{65.2} \\ \hline
\multirow{3}{*}{MT} & P & 66.5 & 79.0 & 66.2 & 78.5 & \textbf{71.0} & \textbf{80.9} \\ \cline{2-8} 
 & R & 68.2 & 78.5 & 68.0 & 77.8 & \textbf{71.8} & \textbf{81.4} \\ \cline{2-8} 
 & F & 67.3 & 78.6 & 67.0 & 78.1 & \textbf{71.2} & \textbf{81.1} \\ \hline
\multirow{3}{*}{PE} & P & 60.9 & 73.2 & 62.3 & 73.2 & \textbf{62.6} & \textbf{74.4} \\ \cline{2-8} 
 & R & 61.2 & 74.0 & 65.8 & 75.1 & \textbf{66.0} & \textbf{75.0} \\ \cline{2-8} 
 & F & 61.1 & 73.6 & 64.0 & 74.1 & \textbf{64.3} & \textbf{74.8} \\ \hline
\multirow{3}{*}{CMV} & P & 54.0 & 65.1 & 55.0 & 68.0 & \textbf{55.7} & \textbf{69.5} \\ \cline{2-8} 
 & R & 53.0 & 62.5 & 59.0 & 65.0 & \textbf{60.0} & \textbf{65.3} \\ \cline{2-8} 
 & F & 53.5 & 63.8 & 57.0 & 66.4 & \textbf{57.8} & \textbf{67.3} \\ \hline
\end{tabular}
\caption{\label{results} Table showing the results on four data sets. Each cell contains the Precision (P), Recall (R) and F-score (F) for Claims as well as the  Macro Precision, Recall and F-score for the binary classification.}
\end{table*}

\section{Results and Experiments}
Table \ref{results} show the results on the four data sets. We compare to two baselines.  The numbers in the \textbf{CNN} column are taken directly from the results of the deep learning experiments mentioned in the work of \newcite{essence}. Their deep learning experiments consisted of 4 different models: a) bidirectional LSTM b) LSTM c) CNN initialized with random word embeddings and d) CNN initialized with word2vec. 
In their experiments for MT and PE, a CNN initialized with random word embeddings gave the best results and for WD a CNN with word2vec gave the best results. As CMV is a new data set we experimented with all four models and obtained the best result using a CNN with random initialization. 
The \textbf{Task-Specific LM Fine-Tuning} column contains the results obtained by fine-tuning the language model on each respective dataset while the \textbf{IMHO LM Fine-Tuning} column contains the results from  fine-tuning the language model on IMHO. As in previous work, we report both Claim F1 and Macro F1.

The experiments were carried out in a 10-fold cross-validation setup with fixed splits into training and test data and the F1 scores are averaged over each of the folds.  Each model was run 10 times to account for variance and the results reported in the table are an average of 10 runs.
We use the same hyper-parameters as \newcite{ulmfit} except for a batch size of 32 for MT and 64 for the remaining data sets. The learning rate for classifier fine-tuning is set to 0.0001. We train our classifier for 5 epochs on each data set. 

We obtain statistically significant results ($p  < 0.05$ using Chi Squared Test)  over all CNN models trained only on the task-specific datasets. 
We also find that for all models, IMHO LM Fine-Tuning even performs better than Task-Specific LM Fine-Tuning, and is significantly better for the MT and WD datasets (which both contain very few claims).  For the MT and WD datasets, Task-Specific LM Fine-Tuning actually performs worse than the CNN models.

\section{Qualitative Analysis}
To understand how using the IMHO dataset improved over the CNN and Task-Specific Fine-Tuning settings, we show examples that were incorrectly classified by the two baseline models but correctly classified by the IMHO Fine-Tuning.  We retrieve the most similar example in the IMHO dataset to these misclassified samples according to TF-IDF over unigrams and bigrams.  Table \ref{table:error_analysis} presents the examples labeled by their dataset and the corresponding IMHO example.  We find that the IMHO dataset contains n-grams indicative of claims, e.g. \textit{can be very rewarding}, \textit{should be taken off the market}, and \textit{should intervene}, demonstrating that the IMHO LM Fine-Tuning learns representations of claims based on discriminatory phrases.  In fact, the CMV example is almost an exact paraphrase of the IMHO example, differing only in the phrase \textit{anecdotal evidence} compared to \textit{my anecdotal experience}. At the same time, we find that many of the topics in these datasets occur in the IMHO dataset as well, such as \textit{public schooling} and \textit{licence fees}, suggesting that the language model learns a bias towards topics as well.

\begin{table*}[]
\renewcommand\thetable{4}
\small
\begin{tabular}{|l|l|}
\hline
Dataset & Sentence \\ \hline
MT & If there must be rent increases , there should also be a cap to avoid nasty surprises \\ \hline
MT & \begin{tabular}[c]{@{}l@{}}Video games namely FIFA in my case , can fascinate young people for hours more intensively and emotionally \\than any sport in the world !\end{tabular} \\ \hline
PE & \begin{tabular}[c]{@{}l@{}}Last but not the least , using public transportation is much safer than using private transportation\end{tabular} \\ \hline
PE & \begin{tabular}[c]{@{}l@{}}In a positive point of view , when people without jobs have hand phones that have  access to the internet , they will\\ be able to browse the net for more job opportunities\end{tabular} \\ \hline
CMV & \begin{tabular}[c]{@{}l@{}} Cheating is evidence , that *something* must be wrong \end{tabular} \\ \hline
\end{tabular}
\caption{Sentences which are actually non-claims but predicted as claims by IMHO Fine-Tuning}
\label{table:error_analysis1}
\end{table*}

While empirical results indicate that IMHO Fine-Tuning helps in claim detection, we also investigated whether the language model introduces any bias towards types of claims. To this end, we also evaluated examples classified incorrectly by the model. Table \ref{table:error_analysis1} shows sentences that are predicted to be opinionated claims by our model but are actually non-claims. 
We note that a portion of these misclassified examples  were premises used to back a claim which could be classified correctly given additional context. For instance, the second example from the MT data set in the table backs the claim \textit{It would be fair to make them into an Olympic event} while the first example from the PE data set backs the claim \textit{There is no reason that governments should hesitate to invest in public transportation, a healthy, safe and economical way of transporting}. While discriminatory phrases like \textit{ should } or \textit{must be} and comparative statements like \textit{ much safer than } or \textit{ more ... than any } are often indicative of claims, the lack of context may lead to incorrect classifications. Language modeling with additional context sentences or jointly modeling context (e.g. by predicting relations  between claims and premises) may address these errors.

\begin{table}[]
\renewcommand\thetable{3}
\vspace{-2ex}
\small
\begin{tabular}{|p{1cm}|p{5.5cm}|}
\hline
Dataset & Sentence \\
\hline
WD & I send my daughter to public school but if I could afford to I would definitely send her to a nearby private school and not have to deal with lots of the  problems in public schools.  \\ 
\hline
IMHO & There is no telling that a private school will be better than public, that 's a parents choice, I pulled my kid from private school and went to public school that choice was made because the school we had access to was new and he excellent ratings and it was superior to the private school. \\
\hline
\hline
MT & That's why they should be taken off the market, unless they're unbreakable . \\
\hline
IMHO & Should be taken off the market.  \\
\hline
\hline
MT & The Tv/Radio licence fee can only be required of all citizens/households equally. \\
\hline
IMHO & Radio 4 and Radio 6 music are pretty much worth the licence fee. \\
\hline
\hline
MT & Since, however, in Russia besides gas and oil only propaganda and corruption rule, the EU should intervene right away. \\  
\hline
IMHO & Neither Russia or the EU should intervene in this case \\
\hline
\hline
CMV & Other than anecdotal evidence, I haven't seen anything to support this claim. \\  
\hline
IMHO & I have personally seen no evidence to support this claim, but that's just my anecdotal experience . \\

\hline
PE & However, flourishing tourism in a place can be very rewarding in terms of local economy. \\
\hline
IMHO & It can be very rewarding. \\
\hline
\end{tabular}
\caption{Sentences from each dataset and their nearest neighbor in the IMHO dataset}
\label{table:error_analysis}
\vspace{-5ex}
\end{table}

\section{Conclusion}
We have collected a large dataset of over 5 million self-labeled opinionated claims and validated their utility on a variety of claim detection domains. Second, we demonstrate that by fine-tuning the language model on our IMHO dataset rather than each individual claim dataset, we obtain statistically significant improvement over previous state-of-the-art performance on each of these datasets on claim detection.  Finally, our empirical results and error analysis show that there are features indicative of claims that transfer across data-sets. 

In the future, we plan to expand this work beyond single sentences as the data-set for LM Fine-Tuning used in our experiments consists of sentences containing IM(H)O without additional context. We plan to experiment with  modeling the context sentences from Reddit as well by using models such as BERT \cite{bert}, which perform well on pair classification tasks, as the fine-tuning step rather than ULMFiT.  As BERT pre-training includes a next-sentence prediction task, we expect this model to be effective for modeling argumentative context and to be beneficial for predicting premise or justifications for these claims and the relations between these argumentative components.

\section{Acknowledgements}
The authors thank Johannes Daxenberger for sharing the sentence level data from their previous  experiments and  Smaranda Muresan for her insightful discussions on improving the contributions of the paper as well as thoughtful ideas on how to conduct a good error analysis. The authors also thank Olivia Winn for suggesting the paper title and the anonymous reviewers for helpful comments.

\bibliography{naaclhlt2019}

\begin{thebibliography}{18}
\expandafter\ifx\csname natexlab\endcsname\relax\def\natexlab#1{#1}\fi

\bibitem[{Al-Khatib et~al.(2016)Al-Khatib, Wachsmuth, Hagen, Kohler, and
  Stein}]{khatib}
Khalid Al-Khatib, Henning Wachsmuth, Matthias Hagen, Jonas Kohler, and Benno
  Stein. 2016.
\newblock Domain mining of argumentative text through distant supervision.
\newblock In \emph{15th Conference of the North American Chapter of the
  Association for Computational Linguistics: Human Language Technologies},
  pages 1395–--1404.

\bibitem[{Daxenberger et~al.(2017)Daxenberger, Eger, Habernal, Stab, and
  Gurevych}]{essence}
Johannes Daxenberger, Steffen Eger, Ivan Habernal, Christian Stab, and Iryna
  Gurevych. 2017.
\newblock What is the essence of a claim? cross-domain claim identification.
\newblock In \emph{Proceedings of the 2017 Conference on Empirical Methods in
  Natural Language Processing}, pages 2055--2066.

\bibitem[{Devlin et~al.(2019)Devlin, Chang, Kenton, and Toutanova}]{bert}
Jacob Devlin, Ming-Wei Chang, Lee Kenton, and Kristina Toutanova. 2019.
\newblock Bert: Pre-training of deep bidirectional transformers for language
  understanding.
\newblock In \emph{Proceedings of the 17th Annual Meeting of the North American
  Association for Computational Linguistics}.

\bibitem[{Eger et~al.(2017)Eger, Daxenberger, and Gurevych}]{end2end}
Steffen Eger, Johannes Daxenberger, and Iryna Gurevych. 2017.
\newblock Neural end-to-end learning for computational argumentation mining.
\newblock In \emph{In Proceedings of the 55th Annual Meeting of the Association
  for Computational Linguistics.}, pages 11–--22.

\bibitem[{Govier(2010)}]{Govier}
Trudy Govier. 2010.
\newblock \emph{A Practical Study of Argument}.
\newblock 7th edition Cengage Learning, Wadsworth.

\bibitem[{Habernal and Gurevych(2017)}]{wd}
Ivan Habernal and Iryna Gurevych. 2017.
\newblock Argumentation mining in user-generated web discourse.
\newblock In \emph{Computational Linguistics, 43(1)}, pages 125–--179.

\bibitem[{Hidey et~al.(2017)Hidey, Musi, Hwang, Muresan, and McKeown}]{cmv}
Christopher Hidey, Elena Musi, Alyssa Hwang, Smaranda Muresan, and Kathleen
  McKeown. 2017.
\newblock Analyzing the semantic types of claims and premises in an online
  persuasive forum.
\newblock In \emph{In Proceedings of the 4th Workshop on Argument Mining.
  EMNLP}, pages 11–--21.

\bibitem[{Howard and Ruder(2018)}]{ulmfit}
Jeremy Howard and Sebastian Ruder. 2018.
\newblock Universal language model fine-tuning for text classification.
\newblock In \emph{Proceedings of the 56th Annual Meeting of the Association
  for Computational Linguistics (Long Papers)}, pages 328--339.

\bibitem[{Merity et~al.(2017)Merity, Xiong, Bradbury, and Socher}]{merity}
Stephen Merity, Caiming Xiong, James Bradbury, and Richard Socher. 2017.
\newblock In proceedings of the international conference on learning
  representations.

\bibitem[{Peldszus and Stede(2016)}]{micro}
Andreas Peldszus and Manfred Stede. 2016.
\newblock An annotated corpus of argumentative microtexts.
\newblock In \emph{In Argumentation and Reasoned Action: Proceedings of the 1st
  European Conference on Argumentation}, pages 810–--815.

\bibitem[{Persing and Ng(2016)}]{essay}
Isaac Persing and Vincent Ng. 2016.
\newblock End-to-end argumentation mining in student essays.
\newblock In \emph{In Proceedings of the 15th Conference of the North American
  Chapter of the Association for Computational Linguistics: Human Language
  Technologies.}, pages 1384--1394.

\bibitem[{Peters et~al.(2018)Peters, Neumann, Iyyer, Gardner, Clark, Lee, and
  Zettlemoyer}]{Peters:2018}
Matthew~E. Peters, Mark Neumann, Mohit Iyyer, Matt Gardner, Christopher Clark,
  Kenton Lee, and Luke Zettlemoyer. 2018.
\newblock Deep contextualized word representations.
\newblock In \emph{Proc. of NAACL}.

\bibitem[{Radford et~al.(2018)Radford, Narasimhan, Salimans, and
  Sutskever}]{openai}
Alec Radford, Karthik Narasimhan, Tim Salimans, and Ilya Sutskever. 2018.
\newblock Improving language understanding by generative pre-training.

\bibitem[{Rosenthal and McKeown(2012)}]{sara}
Sara Rosenthal and Kathy McKeown. 2012.
\newblock Detecting opinionated claims in online discussion.
\newblock In \emph{Inproceedings of the Sixth IEEE International Conference on
  Semantic Computing (IEEE ICSC2012) Special Session on Semantics and
  Sociolinguistics in Social Medi}.

\bibitem[{Schulz et~al.(2018)Schulz, Eger, Daxenberger, Kahse, and
  Gurevych}]{multi}
Claudia Schulz, Steffen Eger, Johannes Daxenberger, Tobias Kahse, and Iryna
  Gurevych. 2018.
\newblock Multi-task learning for argumentation mining in low-resource
  settings.
\newblock In \emph{Proceedings of NAACL-HLT 2018}, pages 35–--41.

\bibitem[{Stab and Gurevych(2017)}]{pe}
Christian Stab and Iryna Gurevych. 2017.
\newblock Parsing argumentation structures in persuasive essays.
\newblock In \emph{Computational Linguistics, pages in press, preprint
  available at arXiv:1604.07370}.

\bibitem[{Toulmin(2003)}]{Toulmin}
Stephen~E. Toulmin. 2003.
\newblock \emph{The Uses of Argument}.
\newblock Cambridge University Press, New York.

\bibitem[{Yang et~al.(2018)Yang, Zhao, Dhingra, He, Cohen, Salakhutdinov, and
  LeCun}]{glomo}
Zhilin Yang, Jake Zhao, Bhuwan Dhingra, Kaiming He, William~W. Cohen, Ruslan
  Salakhutdinov, and Yann LeCun. 2018.
\newblock Glomo: Unsupervisedly learned relational graphs as transferable.
\newblock In \emph{arXiv:1806.05662}.

\end{thebibliography}
\bibliographystyle{acl_natbib}
\end{document}